\documentclass[runningheads]{llncs}

\usepackage{wrapfig}  
\usepackage{caption}
\usepackage{booktabs}  
\usepackage[T1]{fontenc}
\usepackage{graphicx}
\usepackage{marvosym}
\usepackage{color, xcolor}
\usepackage{hyperref}
\usepackage{graphicx}
\usepackage{wrapfig}

\begin{document}
\title{RefineSeg: Dual Coarse-to-Fine Learning for Medical Image Segmentation}
\titlerunning{RefineSeg}
%
\author{Anghong Du\inst{1}
\and Nay Aung\inst{4,5}
\and Theodoros N. Arvanitis\inst{1}
\and Stefan K. Piechnik\inst{2}
\and Joao A C Lima\inst{3}
\and Steffen E. Petersen\inst{4,5}
\and Le Zhang\inst{1,4}\textsuperscript{(\Letter)}
}


\authorrunning{A. Du et al.}
\institute{School of Engineering, College of Engineering and Physical Sciences,\\ University of Birmingham, Birmingham, UK \\
\and Oxford Center for Clinical Magnetic Resonance Research (OCMR),\\ Division of Cardiovascular Medicine, John Radcliffe Hospital,\\University of Oxford, Oxford, UK\\
\and Division of Cardiology, Johns Hopkins University School of Medicine, \\Baltimore, Maryland, USA\\
\and William Harvey Research Institute, NIHR Barts Biomedical Research Centre, Queen Mary University London, London, UK\\
\and Barts Heart Centre, St Bartholomew’s Hospital, Barts Health NHS Trust, West Smithfield, London, UK\\
     \email{axd1038@student.bham.ac.uk; l.zhang.16@bham.ac.uk}
}

\maketitle              
\begin{abstract}
High-quality pixel-level annotations of medical images are essential for supervised segmentation tasks, but obtaining such annotations is costly and requires medical expertise. To address this challenge, we propose a novel coarse-to-fine segmentation framework that relies entirely on coarse-level annotations, encompassing both target and complementary drawings, despite their inherent noise. 
The framework works by introducing transition matrices in order to model the inaccurate and incomplete regions in the coarse annotations. By jointly training on multiple sets of coarse annotations, it progressively refines the network’s outputs and infers the true segmentation distribution, achieving a robust approximation of precise labels through matrix-based modeling. To validate the flexibility and effectiveness of the proposed method, we demonstrate the results on two public cardiac imaging datasets, ACDC and MSCMRseg, and further evaluate its performance on the UK Biobank dataset. Experimental results indicate that our approach surpasses the state-of-the-art weakly supervised methods and closely matches the fully supervised approach. 
\textit{Our code is available at } \url{https://github.com/AnghongDu/RefineSeg-MICCAI2025}.

\end{abstract}

\section{Introduction}


The success of deep supervised learning in image segmentation has been largely attributed to the availability of large-scale datasets with accurate pixel-level annotations~\cite{zhang2025diffuseg} \cite{deng2025kpis}. However, such annotations are especially costly and time-consuming to acquire in the medical domain, 
where expert-level annotation is required and often affected by ambiguous boundaries and inter-observer variability~\cite{gao2023bayeseg} \cite{Zhangneurips}.
These challenges are further compounded by strict privacy regulations, making large-scale, high-quality annotation even more difficult.
For instance, even for experienced experts, precisely delineating cardiac structures such as the left ventricle (LV), right ventricle (RV), and myocardium (MYO) is highly challenging due to ambiguous boundaries. These inherent annotation uncertainties introduce label noise, making the dataset prone to inconsistencies \cite{Zhangneurips}. As a result, despite large imaging repositories like UK Biobank \cite{ukbb}, curating high-quality labels remains a significant challenge, motivating the development of robust learning approaches capable of handling coarse annotations, which is particularly crucial in the medical domain.

To address this challenge, semi-supervised learning (SSL) and weakly supervised learning (WSL) have been widely explored. SSL leverages a small number of labeled samples alongside a large pool of unlabeled data for joint training \cite{pointwssis}. While SSL approaches have demonstrated effectiveness in improving model performance, they still require a considerable amount of fully labeled images as supervision. 
WSL exploit annotations that are easier to obtain than pixel-wise labels, such as bounding boxes \cite{weakpolyp} \cite{boxinst}, scribbles \cite{scribf} \cite{cyclemix} and point labels \cite{pa}.
Despite their lower annotation cost, WSL suffers from annotation noise, as weak labels often fail to provide precise object boundaries, leading to increased uncertainty during training.  
For example, in datasets like ACDC, the MYO is embedded within the LV, making it difficult for bounding boxes to isolate the classes precisely.
A more effective weak annotation strategy is scribble-based annotation, where annotators simply draw lines and circles within the object of interest (OOI) region to provide guidance. However, such methods often rely on post-processing (e.g., ScribFormer~\cite{scribf} uses random walk propagation that assumes closed-loop strokes) to generate full segmentation masks.
In this work, we adopt the coarse annotations that offer more information than scribble labels while avoiding non-target pixels being grabbed into the bounding box. 
Meanwhile, creating coarse annotations, such as rough OOI and non-OOI boundaries, has a cost similar to that of scribble annotations and can be performed by non-experts. Given these advantages, leveraging computational methods to refine noisy pixels from coarse annotations provides a highly efficient and low-cost approach to enriching large-scale dataset annotations.

\textbf{Our contribution:} We propose a novel weakly supervised segmentation framework that enables end-to-end joint training using both \textit{Positive} (target) and \textit{Negative} (complementary) coarse annotations. Unlike the previous WSL approaches, our method models and disentangles the complex mappings from the input images to the coarse annotations and to the true segmentation distribution simultaneously by introducing transition matrices regularizing. For evaluation, we conduct comprehensive experiments on the ACDC, MSCMRseg, and UK Biobank datasets, achieving segmentation performance that surpasses state-of-the-art weakly supervised methods and closely matches fully supervised approach. Moreover, our method offers a promising pathway for making it feasible to train large medical segmentation models, e.g., MedSAM \cite{medsam}, with minimal manual labeling effort while maintaining high performance. 

\section{Method}

\subsection{\ProblemSet}

\begin{figure}[!t]
    \centering
    \includegraphics[width=\linewidth]{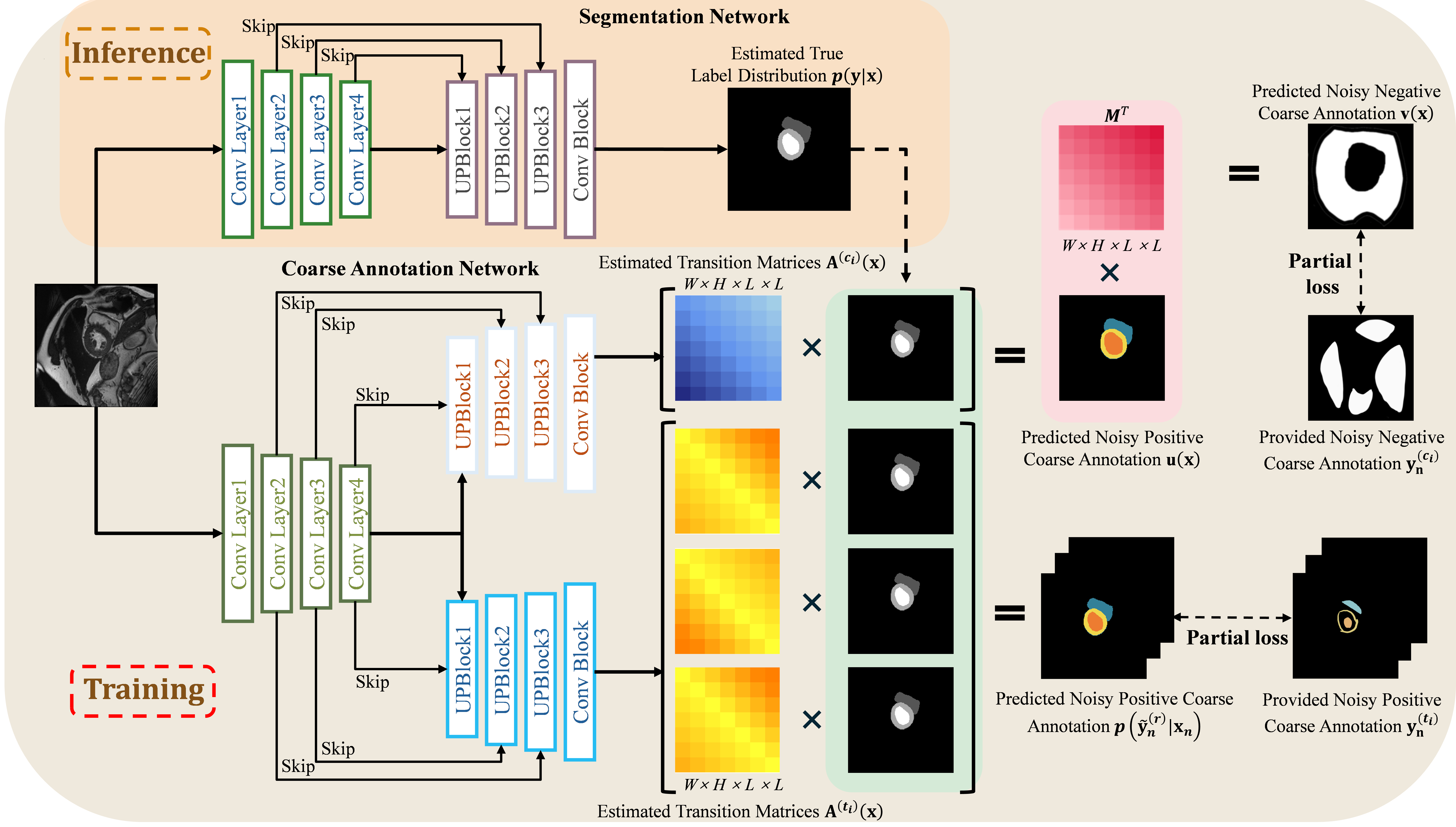}
    \caption{Overview of our proposed coarse-to-fine segmentation framework that jointly learns from \textit{positive} and \textit{negative} coarse annotations.}
    \label{fig:model_pipeline}
\end{figure}

In this work, we address the scenario where a set of images 
$
\{\mathbf{x}_n \in \mathds{R}^{W \times H \times C} \}_{n=1}^{N}
$
(with \(W, H, C\) denoting the width, height and channels of the image) are assigned \textbf{Positive} (target) and \textbf{Negative} (complementary) coarse labels
$
\{ \mathbf{y}_n^{(t_i)}, \mathbf{y}_n^{(c_i)} \in Y^{W \times H} \}_{i=1}^{P},\quad n=1,\dots,N.
$
Here, \( P \) denotes the total number of annotation strategies, \( N \) represents the total number of images in the dataset, and \( Y = \{1,2,\dots,L\} \) denotes the set of possible classes.
Figure~\ref{fig:model_pipeline} illustrates our proposed end-to-end joint training framework.
Our problem can be formulated as estimating the unobserved true segmentation distribution \( \textit{p}(\mathbf{y}_n | \mathbf{x}_n) \) from the dataset $\mathcal{D} = \{\mathbf{x}_n, ({\mathbf{y}}_n^{(t_i)}, {\mathbf{y}}_n^{(c_i)})\}_{i=1}^{P}$ with multiple coarse drawing labels.


\subsection{Joint Training with Multiple Coarse Annotations}

In this section, we describe how to jointly learn the true segmentation distribution \( p(\mathbf{y}_n | \mathbf{x}_n) \) alongside the transition matrix \( A^{(t_i)} \) and \( A^{(c_i)} \) from multiple coarse annotation networks. In short, we minimize the joint training loss functions of the probability model using the observed positive and negative coarse labels. A detailed description is provided below.

\textbf{Pixel-wise Transition Matrix.}
Different from traditional methods that assume all images share a same transition matrix \cite{hoogeboom2021argmax}, our approach leverages the independent pixel-wise transition matrix \cite{Zhangneurips} to refine segmentation predictions for each input image. Our transition matrix is built on the \textit{Markov chain transition assumption} \cite{markov}, which ensures the current segmentation state depends only on its immediate previous state. 
In particular, we refer to the \( L \times L \) matrix, where each \((m,k)\)-th element is defined by :
$ A^{(r)}(\mathbf{x}, u, v)_{mk} := p(\tilde{\mathbf{y}}^{(r)}_{uv} = m \mid \mathbf{y}_{uv} = k, \mathbf{x})
, \quad \forall m,k \in \{1, \dots, L\},
$ as the transition matrix at pixel \( (u, v) \) in image \( \mathbf{x} \),  \( r \in \{t_i, c_i\} \) represents the coarse annotation strategies.

Given an image \( \mathbf{x}_n \), under the assumption that annotations at different pixels are conditionally independent, the probability of the observed coarse labels on each pixel \( (u,v) \) can be formulated as:

\begin{small}
\begin{equation}
    p(\tilde{\mathbf{y}}_n^{(r)}(u,v) | \mathbf{x}_n) 
    = \sum_{\mathbf{y}_n \in Y} A^{(r)}(u,v) \ \cdot p(\mathbf{y}_n (u,v)| \mathbf{x}_n),
    \label{eq:probabilistic transition process}
\end{equation}
\end{small}\\
where \( p(\mathbf{y}_n (u,v) | \mathbf{x}_n) \) represents the predicted fine-grained label distribution, and \( p(\tilde{\mathbf{y}}_n^{(r)}(u,v) | \mathbf{x}_n) \) represents the predicted coarse label distribution. Annotation network estimates the pixel-wise transition matrices 
\(\left\{ {A}^{(r)}(\mathbf{x}) \in [0,1]^{W \times H \times L \times L} 
\right\}_{\scriptscriptstyle r=1}^{\scriptscriptstyle P} \)
for input image \( \mathbf{x} \). 
Equation~\eqref{eq:probabilistic transition process} describes the probabilistic transition process in which annotation network \( r \) adjusts \( p(\mathbf{y}_n (u,v) | \mathbf{x}_n) \) to align with the coarse labels.

\textbf{Learning with positive coarse label.}
Given a training input \( \mathbf{x}_n \) and a positive coarse label \( {\mathbf{y}}_n^{(t_i)} \), we optimize the transition matrix \( A^{(t_i)} \) of the coarse annotation network by minimizing the following hybrid loss function: 


\begin{equation}
    \mathcal{L}_{pos}^{(t_i)} = \sum_{i=1}^{P_{pos}} \left[ \alpha_{i} \mathcal{L}_{ce}^{(t_i)}
    + \beta_{i} \mathcal{L}_{dice}^{(t_i)} \right],
    \label{eq:pos_loss}
\end{equation}
where \( \mathcal{L}_{ce}^{(t_i)} \) and \( \mathcal{L}_{dice}^{(t_i)} \) denote the \textit{cross-entropy loss} and \textit{Dice loss}, which together form the hybrid loss. \( P_{pos} \) represents the set of positive annotation strategies within \( P \). \( \alpha_{i} \) and \( \beta_{i} \) are weight parameters that balance the contribution of each loss term.
Minimizing Equation~\eqref{eq:pos_loss} encourages the transition matrix \( A^{(t_i)} \) adjusted segmentation probability map \(p(\mathbf{y}_n | \mathbf{x}_n) \) to align closely with the provided positive coarse label
 \( {\mathbf{y}}_n^{(t_i)} \). However, directly applying  \( \mathcal{L}_{ce}^{(t_i)} \) and \( \mathcal{L}_{dice}^{(t_i)} \) to the entire image is ineffective due to severe class imbalance. The CE loss can be minimized by predicting all pixels as the most frequent background class. Although Dice loss mitigates class imbalance, annotation noise in coarse labels makes unannotated background regions unreliable, as they may still contain target information.

To address this issue, \cite{partial} proposed to restrict loss computation to only the annotated pixels while ignoring unverified regions. Building on this idea, we design the  partial loss function, formulated as:

\begin{equation}
    \mathcal{L}_{ce}^{(t_i)} = - \frac{1}{|\mathcal{R}_{pos}^{(t_i)}|} 
    \sum_{(u,v) \in \mathcal{R}_{pos}^{(t_i)}} 
    {\mathbf{y}}_n^{(t_i)}(u,v) 
    \log \left[A^{(t_i)} p(\mathbf{y}_n | \mathbf{x}_n) (u,v)\right],
    \label{eq:partial_ce_loss}
\end{equation}

\begin{equation}
    \mathcal{L}_{dice}^{(t_i)} = 1 - \frac{2 \sum_{(u,v) \in \mathcal{R}_{pos}^{(t_i)}} A^{(t_i)} p(\mathbf{y}_n | \mathbf{x}_n) (u,v) {\mathbf{y}}_n^{(t_i)}(u,v)}
    {\sum_{(u,v) \in \mathcal{R}_{pos}^{(t_i)}} \Big( A^{(t_i)} p(\mathbf{y}_n | \mathbf{x}_n) (u,v) + {\mathbf{y}}_n^{(t_i)}(u,v) \Big) },
    \label{eq:partial_dice_loss}
\end{equation}
where \( \mathcal{R}_{pos}^{(t_i)} \) represents the set of pixels labeled as positive in \({\mathbf{y}}_n^{(t_i)} \), excluding negative pixels.  

\textbf{Learning with negative coarse label.}
For some situations, it is easier to provide negative  coarse labels to help the model predict the true label distribution. However, if we directly apply loss, as in Equation~\eqref{eq:pos_loss}, when learning with these negative coarse labels, the model can only learn a mapping \( \mathds{R} \to Y \) that attempts to predict the conditional probability \( p(\tilde{\mathbf{y}}_n^{(c_i)} | \mathbf{x}_n) \) and the corresponding negative pixel that predicts \( \tilde{\mathbf{y}}_n^{(c_i)}(u,v) \) for a input image \( \mathbf{x}_n \).

To address this issue, inspired by \cite{complementary} \cite{Zhangneurips}, which summarizes all the probabilities into a transition matrix \( M \in \mathds{R}^{L \times L} \), where
$
M_{mk}(u,v) = p(\tilde{\mathbf{y}}_n^{(c)}(u,v) = m | \mathbf{y}_n(u,v) = k, \mathbf{x}_n)
$
and 
$
M_{mm}(u,v) = 0
$,
we introduce a transition-based negative learning approach. Here, \( M_{mk} \) denotes the entry in the \( m \)-th row and \( k \)-th column of \( M \), representing the probability of flipping the true label \( k \) into the complementary label \( m \). 
We achieve this by introducing a linear transformation layer in the negative coarse label learning channel. This layer outputs \( \mathbf{v}(\mathbf{x}_n) \) by multiplying the output of the coarse annotation network \( A^{(c_i)} p(\mathbf{y}_n | \mathbf{x}_n) \), denoted as \( \mathbf{u}(\mathbf{x}_n) \), with the transposed transition matrix \( M^\top \). 

We also observe that \( p(\tilde{\mathbf{y}}^{(c)} | \mathbf{x}) \) can be transformed into \( p(\tilde{\mathbf{y}}^{(t)} | \mathbf{x}) \) using the transition matrix \( M \),
\begin{equation}
\begin{aligned}
    p(\tilde{\mathbf{y}}^{(t)}_{uv} = k \mid \mathbf{x}_n) 
    &= \sum_{m \neq k} p(\tilde{\mathbf{y}}^{(t)}_{uv} = k \mid \tilde{\mathbf{y}}^{(c)}_{uv} = m, \mathbf{x}_n) p(\tilde{\mathbf{y}}^{(c)}_{uv} = m \mid \mathbf{x}_n). \\
\end{aligned}
\label{eq:transition_from_c_to_t}
\end{equation}


To enable end-to-end learning rather than transferring after training, we define:
\begin{equation}
    \mathbf{v}(\mathbf{x}_n) = M^\top \mathbf{u}(\mathbf{x}_n).
\end{equation}

Here, we apply the transposed transition matrix \( M^\top \) to ensure that the learned distribution \( \mathbf{v}(\mathbf{x}_n) \) aligns with the negative coarse label. Then,
$
\mathcal{L}_{neg} =  
    \sum_{i=1}^{P_{neg}}
    (M^\top(A^{(c_i)} p(\mathbf{y}_n | \mathbf{x}_n)), {\mathbf{y}}_n^{(c_i)})
$, where \( P_{neg} \) represents the set of positive annotation strategies within \( P \).

\textbf{Regularizing for the Transition Matrices.}
In the joint training process, the model may rely excessively on the transition matrix to adjust \( p(\mathbf{y}_n | \mathbf{x}_n) \) to fit the coarse labels rather than making subtle refinements to enhance the quality of the segmentation distribution. Existing studies~\cite{Zhangneurips} employ trace constraints to mitigate overfitting to coarse labels. However, negative coarse labels differ significantly from positive labels, making trace constraints insufficient to prevent unstable optimization caused by annotation heterogeneity. To address this, we propose a identity matrix regularization term to stabilize the transition matrices:
\begin{equation}
    \mathcal{L}_{reg} = \sum_{i=1}^{P_{pos}} || A^{(t_i)} - I ||_F^2,
    \label{eq:unit_matrix_regularization}
\end{equation}
where \( I \) is the identity matrix, \( || \cdot ||_F^2 \) denotes the Frobenius norm. This regularization preserves the structural integrity of segmentation predictions while allowing necessary refinements, preventing transition matrices from learning trivial mappings that overfit coarse labels instead of capturing meaningful features.

Finally, we combine the positive annotation loss $\mathcal{L}_{pos}$ and the negative annotation loss $\mathcal{L}_{neg}$ as our objective and optimize the following:

\begin{equation}
    \mathcal{L}_{total} = \sum_{i=1}^{P_{pos}} \mathcal{L}_{pos}(A^{(t_i)} p(\mathbf{y}_n | \mathbf{x}_n), {\mathbf{y}}_n^{(t_i)}) + 
    \sum_{i=1}^{P_{neg}}
    \mathcal{L}_{neg}(M^\top(A^{(c_i)} p(\mathbf{y}_n | \mathbf{x}_n)), {\mathbf{y}}_n^{(c_i)}) +  \lambda \mathcal{L}_{reg},
    \label{eq:final hybrid loss}
\end{equation}
where \( \lambda \) is weight parameter for the regularization term.

\section{Experiments}
\textbf{Dataset.} Two public cardiac datasets, ACDC \cite{ACDC} and MSCMRseg \cite{MSCMR1} \cite{MSCMR2}, along with UK Biobank (UKBB) \cite{ukbb} dataset are adopted to evaluate our method. ACDC contains cine MRI scans of 100 patients, MSCMRseg includes LGE MRI scans of 45 cardiomyopathy patients, and UKBB cardiac dataset comprises short-axis CMR images from 600 subjects. All three datasets are provided with ground-truth annotations meticulously performed by experienced cardiovascular imaging specialists. 
To obtain positive-negative coarse annotations, we erode the available segmentation masks for ACDC and MSCMRseg datasets, following the approach in \cite{simu}. 
For UKBB, we obtain the realistic coarse annotations by manually annotating the data following the principles in
\cite{getscribble}. We also follow the approaches in \cite{pa}, \cite{weakpolyp} and \cite{boxinst} to obtain the box and point annotations on ACDC and MSCMRseg datasets, and obtain scribbles on UKBB dataset for the comparison experiments.
Across all datasets, we uniformly use 80\% of each dataset for training and 20\% for testing. Note that, during training, only coarse annotations are adopted in our framework.

\textbf{Implementation Settings.}
Our framework was implemented in PyTorch and employed the 2D U-Net \cite{unet} as the network architecture. For all datasets, we resized or padded all images to a uniform size of 224 $\times$ 224 pixels. 
For data augmentation, zero-mean and unit-variance normalization, random flipping, and random rotation were applied. 
Before being input to the model, each image is normalized using min-max scaling to bring pixel values into the \( [0,1] \) range. 
The optimizer used was AdamW, with an initial learning rate of 1e-3 and a weight decay of 2e-5. 
In Equation~\eqref{eq:pos_loss}, we empirically set \( \alpha = 0.6 \) and \( \beta = 0.4 \). For Equation~\eqref{eq:final hybrid loss}, we set \( \lambda = 0.2 \). 
All models were trained using one single NVIDIA A100 40GB GPU for 200 epochs.

\textbf{Baseline settings and Evaluation metrics.}
We conduct our experiments under the assumption that no ground truth labels are available. Specifically, we compare multiple weakly supervised approaches that leverage scribble annotations \cite{cyclemix} \cite{scribf}, coarse annotations \cite{lcmil}, box-level annotations \cite{boxinst} \cite{weakpolyp}, and point labels \cite{pa}. Additionally, we include a semi-supervised method \cite{pointwssis} that utilizes point annotations, alongside the fully supervised nnU-Net \cite{nnunet}, which represents the state-of-the-art in cardiac segmentation when trained with ground truth labels. To evaluate segmentation performance, we use the Dice score for RV, LV, and MYO. In addition, we report the average Dice across these three regions to provide a comprehensive measure of segmentation accuracy.

\begin{table}[!t]
    \setlength{\tabcolsep}{1mm}
    \renewcommand{\arraystretch}{0.9}
    \centering
    \caption{The results of Dice score on ACDC and MSCMRseg datasets. Bold denotes the best performance among all methods except nnUNet. Numbers in bold indicate the best method that statistically ($p <$ 0.01) better than other methods by computing the $p$ values of paired $t$-tests on Dice score.}
    \begin{tabular}{l|c|cccc|cccc}
        \toprule
        \multirow{2}{*}{\textbf{Methods}} & \multirow{2}{*}{\textbf{Annotations}} &
        \multicolumn{4}{c|}{ACDC} & \multicolumn{4}{c}{MSCMRseg} \\
        \cmidrule(lr){3-6}\cmidrule(lr){7-10}
         & & LV & MYO & RV & Avg & LV & MYO & RV & Avg \\
        \midrule
        
        \multicolumn{10}{l}{\textbf{Weakly-supervised}} \\
        \midrule
        PA-Seg \cite{pa} & points
            & .841 & .723 & .729 & .764 
            & .771 & .609 & .534 & .638 \\

        WeakPolyp \cite{weakpolyp} & box
            & .836 & .718 & .632 & .728
            & .767 & .566 & .516 & .615 \\

        BoxInst \cite{boxinst} & box
            & .803 & .674 & .583 & .686
            & .747 & .503 & .494 & .581 \\

        ScribFormer \cite{scribf} & scribbles
            & .922 & .871 & .871 & .888
            & .896 & .813 & .807 & .839 \\

        CycleMix \cite{cyclemix} & scribbles
            & .883 & .798 & .863 & .848 
            & .870 & .739 & .791 & .800 \\

        LC-MIL \cite{lcmil}  & coarse
            & .873 & .684 & .561 & .706
            & .723 & .537 & .520 & .593 \\

        \textbf{Ours} & coarse & \textbf{.938} & \textbf{.884} & \textbf{.881} & \textbf{.901} 
                      & \textbf{.904} & \textbf{.839} & \textbf{.812} & \textbf{.852} \\
        
        \midrule
        \multicolumn{10}{l}{\textbf{Semi-supervised}} \\
        \midrule
        PointWSSIS \cite{pointwssis} & 5\%mask+point
            & .901 & .777 & .807 & .828
            & .844 & .748 & .705 & .765 \\
        
        \midrule
        \multicolumn{10}{l}{\textbf{Fully supervised}} \\
        \midrule
        nnUNet \cite{nnunet} & mask
            & .943 & .901 & .915 & .920
            & .944 & .882 & .880 & .902 \\
       
        \bottomrule
    \end{tabular}
    \label{tab:ACDC and MSCMR}
\end{table}

\begin{figure*}[!t]
    \centering
    \includegraphics[width=\textwidth]{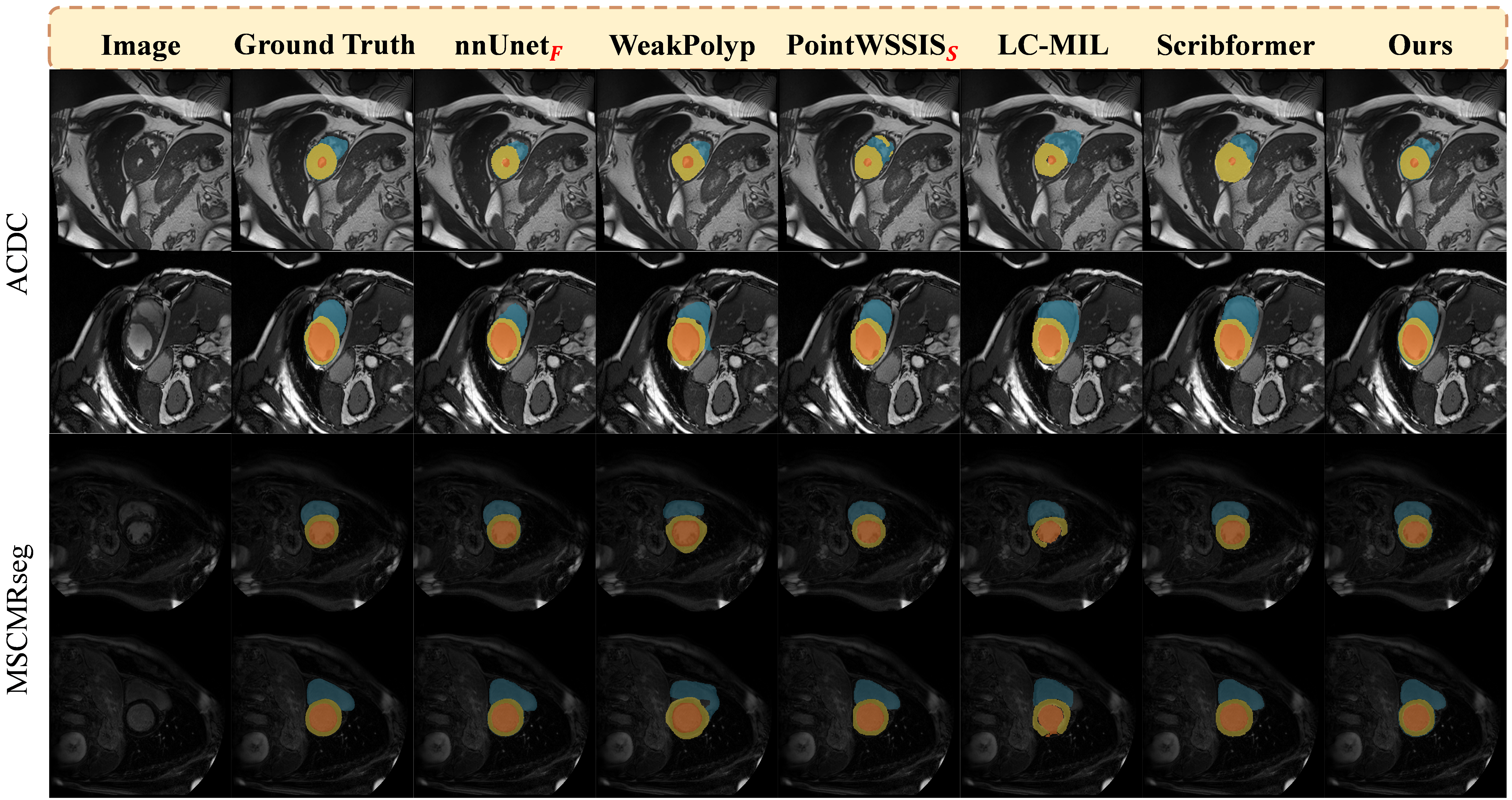} 
    \caption{Qualitative comparisons of our framework with state-of-the-art weakly supervised methods on the ACDC and MSCMRseg datasets. Subscripts \textcolor{red}{\textit{F}} and \textcolor{red}{\textit{S}} indicate segmentation models trained with full supervision and semi-supervision.}
    \label{fig:qualitative_comparison}
\end{figure*}

\begin{figure}[!t]
    \centering
    \begin{minipage}{0.6\textwidth}
        \includegraphics[width=\textwidth]{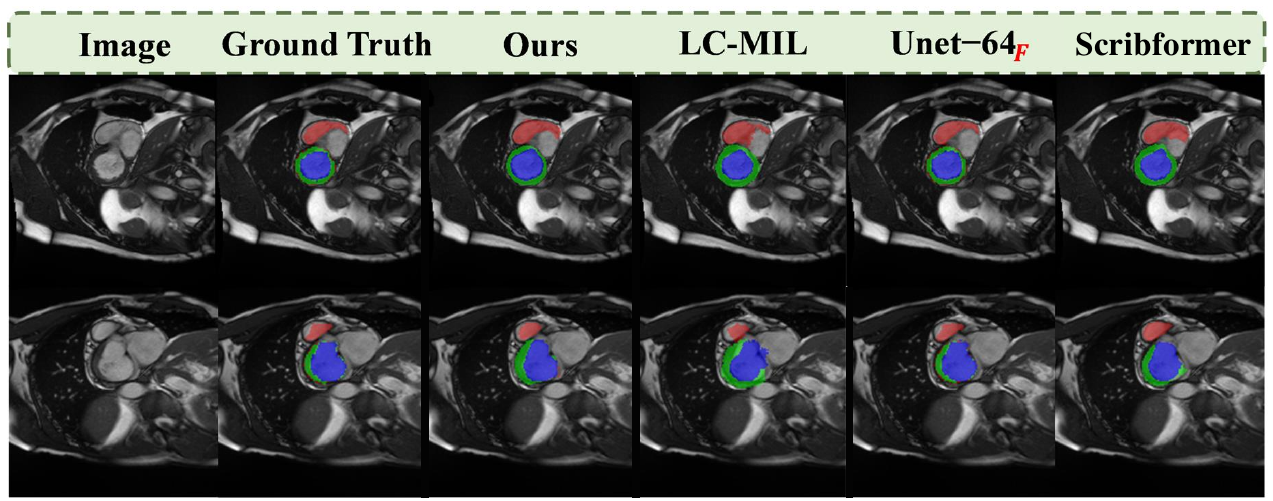}
        \caption{Visualization of the segmentation performance of different supervision strategies on UKBB datasets. Subscripts \textcolor{red}{\textit{F}} denote segmentation models trained with full supervision.}\label{fig:qualitative_comparison2}
    \end{minipage}
    \hfill
    \begin{minipage}{0.35\textwidth}
        \includegraphics[width=\textwidth]{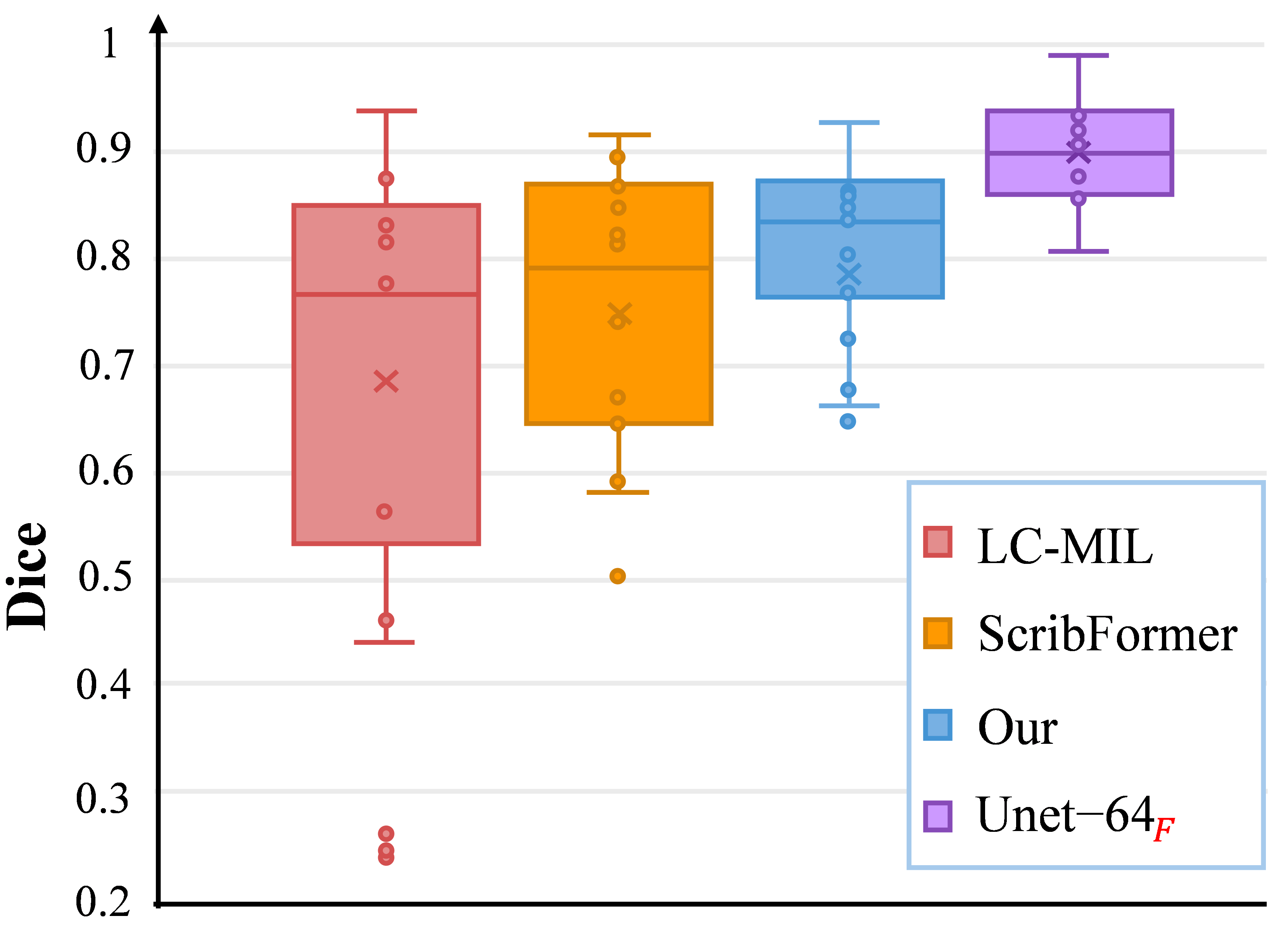}
        \caption{The Dice distribution of different supervision strategies on the UKBB dataset.}\label{fig:ukbb_comparison2}
    \end{minipage}
\end{figure}

\textbf{Quantitative Comparison.} Table~\ref{tab:ACDC and MSCMR} compares the Dice performance of models trained with different supervision strategies and backbones on the ACDC and MSCMRseg datasets, including results reported in \cite{scribf} and \cite{cyclemix}. Our proposed framework outperforms multiple weakly supervised approaches. 
Specifically, in the first section, our method surpasses the ScribFormer \cite{scribf} by 1.4\% in average Dice on ACDC dataset (0.902 vs. 0.888) and by 1.3\% on MSCMRseg dataset (0.852 vs. 0.839). 
The second and third sections of Table~\ref{tab:ACDC and MSCMR} further present the comparison of our framework with semi-supervised \cite{pointwssis} and fully-supervised \cite{nnunet} methods. Results indicate that our approach outperforms semi-supervised training approach with partial masks and achieves performance close to full supervision approach.
This demonstrates the effectiveness of jointly training with both \textit{positive} and \textit{negative} coarse annotations. The inclusion of negative coarse annotations further enhances the model’s ability to extract positive features while imposing stronger regularization, leading to more robust feature representation. Figure~\ref{fig:ukbb_comparison2} highlights our model not only achieves a higher average Dice score but also exhibits greater performance stability compared to other weakly supervised approaches. Notably, our study is the first weakly supervised benchmark compared to fully supervised approach on the UKBB dataset, providing a valuable reference for future research in this area.



\textbf{Qualitative Comparison.}
Figure~\ref{fig:qualitative_comparison} presents the segmentation visualization of different methods on the ACDC and MSCMRseg datasets, and Figure~\ref{fig:qualitative_comparison2} shows the performance of the approaches with different supervision strategies on UKBB dataset. 
As shown in Figure~\ref{fig:qualitative_comparison}, both WeakPolyp and LC-MIL struggle to accurately preserve structural shape. In contrast, our approach effectively integrates and optimizes features from different coarse annotations, resulting in a more comprehensive representation. This capability mitigates the inherent limitations of Unet, which tends to focus primarily on localized features.
Moreover, compared to the weakly supervised results shown in Figure~\ref{fig:qualitative_comparison2}, our approach not only produces results that are closer to the ground truth but also maintains shape integrity comparable to fully supervised method.


\section{Conclusion}
Pixel-level annotation remains a major challenge in medical image segmentation, constraining further progress in this field. To address this challenge, we propose a novel weakly supervised segmentation framework that enables end-to-end training using \textit{positive-negative} coarse annotations by introducing transition matrices. Experimental results show our method outperforms state-of-the-art weakly supervised approaches and closely matches fully supervised model. By reducing annotation costs without compromising performance, our method enhances efficiency and underscores the potential of weakly supervised learning for cost-effective, high-precision segmentation. Furthermore, it presents a promising approach for facilitating the training of large medical segmentation models, including MedSAM \cite{medsam}.

\section*{Disclosure of Interests}
The authors declare that there are no competing interests related to this work.

%
%
%
%



\bibliographystyle{splncs04}  
\bibliography{Paper-0555}    

\begin{thebibliography}{10}
\providecommand{\url}[1]{\texttt{#1}}
\providecommand{\urlprefix}{URL }
\providecommand{\doi}[1]{https://doi.org/#1}

\bibitem{ACDC}
Bernard, O., Lalande, A., Zotti, C., Cervenansky, F., Yang, X., Heng, P.A., Cetin, I., Lekadir, K., Camara, O., Ballester, M.A.G., et~al.: Deep learning techniques for automatic mri cardiac multi-structures segmentation and diagnosis: is the problem solved? IEEE transactions on medical imaging  \textbf{37}(11),  2514--2525 (2018)

\bibitem{simu}
Castro, D.C., Tan, J., Kainz, B., Konukoglu, E., Glocker, B.: Morpho-mnist: Quantitative assessment and diagnostics for representation learning. Journal of Machine Learning Research  \textbf{20}(178),  1--29 (2019)

\bibitem{deng2025kpis}
Deng, R., Yao, T., Tang, Y., Guo, J., Lu, S., Xiong, J., Yu, L., Cap, Q.H., Cai, P., Lan, L., et~al.: Kpis 2024 challenge: Advancing glomerular segmentation from patch-to slide-level. Medical Image Analysis  (2025)

\bibitem{markov}
Gagniuc, P.A.: Markov chains: from theory to implementation and experimentation. John Wiley \& Sons (2017)

\bibitem{gao2023bayeseg}
Gao, S., Zhou, H., Gao, Y., Zhuang, X.: Bayeseg: Bayesian modeling for medical image segmentation with interpretable generalizability. Medical Image Analysis  \textbf{89},  102889 (2023)

\bibitem{MSCMR2}
Gao, S., Zhou, H., Gao, Y., Zhuang, X.: Bayeseg: Bayesian modeling for medical image segmentation with interpretable generalizability. Medical Image Analysis  \textbf{89},  102889 (2023)

\bibitem{hoogeboom2021argmax}
Hoogeboom, E., Nielsen, D., Jaini, P., Forr{\'e}, P., Welling, M.: Argmax flows and multinomial diffusion: Learning categorical distributions. Advances in Neural Information Processing Systems  \textbf{34},  12454--12465 (2021)

\bibitem{nnunet}
Isensee, F., Jaeger, P.F., Kohl, S.A., Petersen, J., Maier-Hein, K.H.: nnu-net: a self-configuring method for deep learning-based biomedical image segmentation. Nature methods  \textbf{18}(2),  203--211 (2021)

\bibitem{pointwssis}
Kim, B., Jeong, J., Han, D., Hwang, S.J.: The devil is in the points: Weakly semi-supervised instance segmentation via point-guided mask representation. In: Proceedings of the IEEE/CVF Conference on Computer Vision and Pattern Recognition. pp. 11360--11370 (2023)

\bibitem{scribf}
Li, Z., Zheng, Y., Shan, D., Yang, S., Li, Q., Wang, B., Zhang, Y., Hong, Q., Shen, D.: Scribformer: Transformer makes cnn work better for scribble-based medical image segmentation. IEEE Transactions on Medical Imaging  (2024)

\bibitem{medsam}
Ma, J., He, Y., Li, F., Han, L., You, C., Wang, B.: Segment anything in medical images. Nature Communications  \textbf{15}(1), ~654 (2024)

\bibitem{unet}
Ronneberger, O., Fischer, P., Brox, T.: U-net: Convolutional networks for biomedical image segmentation. In: Medical image computing and computer-assisted intervention--MICCAI 2015: 18th international conference, Munich, Germany, October 5-9, 2015, proceedings, part III 18. pp. 234--241. Springer (2015)

\bibitem{ukbb}
Sudlow, C., Gallacher, J., Allen, N., Beral, V., Burton, P., Danesh, J., Downey, P., Elliott, P., Green, J., Landray, M., et~al.: Uk biobank: an open access resource for identifying the causes of a wide range of complex diseases of middle and old age. PLoS medicine  \textbf{12}(3),  e1001779 (2015)

\bibitem{boxinst}
Tian, Z., Shen, C., Wang, X., Chen, H.: Boxinst: High-performance instance segmentation with box annotations. In: Proceedings of the IEEE/CVF Conference on Computer Vision and Pattern Recognition. pp. 5443--5452 (2021)

\bibitem{getscribble}
Valvano, G., Leo, A., Tsaftaris, S.A.: Learning to segment from scribbles using multi-scale adversarial attention gates. IEEE Transactions on Medical Imaging  \textbf{40}(8),  1990--2001 (2021)

\bibitem{lcmil}
Wang, Z., Saoud, C., Wangsiricharoen, S., James, A.W., Popel, A.S., Sulam, J.: Label cleaning multiple instance learning: Refining coarse annotations on single whole-slide images. IEEE transactions on medical imaging  \textbf{41}(12),  3952--3968 (2022)

\bibitem{weakpolyp}
Wei, J., Hu, Y., Cui, S., Zhou, S.K., Li, Z.: Weakpolyp: You only look bounding box for polyp segmentation. In: International Conference on Medical Image Computing and Computer-Assisted Intervention. pp. 757--766. Springer (2023)

\bibitem{partial}
Wen, H., Cui, J., Hang, H., Liu, J., Wang, Y., Lin, Z.: Leveraged weighted loss for partial label learning. In: International conference on machine learning. pp. 11091--11100. PMLR (2021)

\bibitem{complementary}
Yu, X., Liu, T., Gong, M., Tao, D.: Learning with biased complementary labels. In: Proceedings of the European conference on computer vision (ECCV). pp. 68--83 (2018)

\bibitem{pa}
Zhai, S., Wang, G., Luo, X., Yue, Q., Li, K., Zhang, S.: Pa-seg: Learning from point annotations for 3d medical image segmentation using contextual regularization and cross knowledge distillation. IEEE transactions on medical imaging  \textbf{42}(8),  2235--2246 (2023)

\bibitem{cyclemix}
Zhang, K., Zhuang, X.: Cyclemix: A holistic strategy for medical image segmentation from scribble supervision. In: Proceedings of the IEEE/CVF Conference on Computer Vision and Pattern Recognition. pp. 11656--11665 (2022)

\bibitem{Zhangneurips}
Zhang, L., Tanno, R., Xu, M.C., Jin, C., Jacob, J., Cicarrelli, O., Barkhof, F., Alexander, D.: Disentangling human error from ground truth in segmentation of medical images. Advances in Neural Information Processing Systems  \textbf{33},  15750--15762 (2020)

\bibitem{zhang2025diffuseg}
Zhang, L., Wu, F., Bronik, K., Papiez, B.W.: Diffuseg: Domain-driven diffusion for medical image segmentation. IEEE Journal of Biomedical and Health Informatics  (2025)

\bibitem{MSCMR1}
Zhuang, X.: Multivariate mixture model for myocardial segmentation combining multi-source images. IEEE transactions on pattern analysis and machine intelligence  \textbf{41}(12),  2933--2946 (2018)

\end{thebibliography}
\end{document}